\documentclass[letterpaper, 10 pt, conference]{ieeeconf}  % Comment this line out if you need a4paper

\IEEEoverridecommandlockouts                              % This command is only needed if 
\let\labelindent\relax
\usepackage{caption}
\usepackage{enumitem}
\usepackage[table]{xcolor}
\usepackage{multirow}
\usepackage{booktabs}

\newcommand{\ie}{i.e., }
\newcommand{\eg}{e.g., }
\newcommand{\etc}{etc.}
\usepackage[pdftex]{graphicx}
\usepackage{nicefrac}
\usepackage{threeparttable}
\usepackage{tablefootnote}
\usepackage[normalem]{ulem}
\usepackage{blindtext}
\usepackage[absolute]{textpos}
\usepackage{url}
\useunder{\uline}{\ul}{}
\graphicspath{{./images/}}
\DeclareGraphicsExtensions{.pdf,.jpeg,.png}

\title{\LARGE \bf
Understanding and Modeling the Effects of Task and Context on Drivers' Gaze Allocation
}

\author{Iuliia Kotseruba and John K. Tsotsos% <-this % stops a space
\thanks{Both authors are with the Department of Computer Science and Electrical Engineering, York University, Canada.
        Email: {\tt \{yulia84, tsotsos\}@yorku.ca}}%
}

\begin{document}

\begin{textblock}{10}(1,0.5)
\noindent \footnotesize Accepted at IEEE Intelligent Vehicles Symposium (IV), 2024
\end{textblock}

\maketitle
\thispagestyle{empty}
\pagestyle{empty}

%%%%%%%%%%%%%%%%%%%%%%%%%%%%%%%%%%%%%%%%%%%%%%%%%%%%%%%%%%%%%%%%%%%%%%%%%%%%%%%%
\begin{abstract}
To further advance driver monitoring and assistance systems, it is important to understand how drivers allocate their attention, in other words, where do they tend to look and why. Traditionally, factors affecting human visual attention have been divided into bottom-up (involuntary attraction to salient regions) and top-down (driven by the demands of the task being performed). Although both play a role in directing drivers' gaze, most of the existing models for drivers' gaze prediction apply techniques developed for bottom-up saliency and do not consider influences of the drivers' actions explicitly. Likewise, common driving attention benchmarks lack relevant annotations for drivers' actions and the context in which they are performed. Therefore, to enable analysis and modeling of these factors for drivers' gaze prediction, we propose the following: 1) we correct the data processing pipeline used in DR(eye)VE to reduce noise in the recorded gaze data; 2) we then add per-frame labels for driving task and context; 3) we benchmark a number of baseline and SOTA models for saliency and driver gaze prediction and use new annotations to analyze how their performance changes in scenarios involving different tasks; and, lastly, 4) we develop a novel model that modulates drivers' gaze prediction with explicit action and context information. While reducing noise in the DR(eye)VE gaze data improves results of all models, we show that using task information in our proposed model boosts performance even further compared to bottom-up models on the cleaned up data, both overall (by 24\% KLD and 89\% NSS) and on scenarios that involve performing safety-critical maneuvers and crossing intersections (by up to 10--30\% KLD). Extended annotations and code are available at \url{https://github.com/ykotseruba/SCOUT}.
\end{abstract}

%%%%%%%%%%%%%%%%%%%%%%%%%%%%%%%%%%%%%%%%%%%%%%%%%%%%%%%%%%%%%%%%%%%%%%%%%%%%%%%%
\section{Introduction}

Despite being commonplace, driving is inherently risky. Since lapsed or misplaced attention of drivers has been linked to increased odds of traffic accidents \cite{2014_JSR_Caird}, much of the research on driver attention allocation has been safety-centered. In particular, a large body of research is dedicated to identifying and mitigating negative effects of engaging in non-driving related activities (\eg using a cell phone) and investigating hazard anticipation \cite{2021_arXiv_Kotseruba}. 

In comparison, routine decision-making during driving is less studied. An important property of driving is that it is a visuomotor activity, \ie motor actions of drivers are linked to vision, which provides most of the information \cite{1996_Perception_Sivak}. However, drivers do not process the entire scene at once due to the limited field of view and varying acuity within it (foveation) \cite{1935_AO_Osterberg}. Instead, they move their eyes sequentially to bring different elements of the scene into the fovea. Selection of the next location to look has been described in vision science as either \textit{bottom-up} involuntary attraction to salient regions or \textit{top-down} task-driven sampling of areas relevant to current or future action \cite{tsotsos2011computational, nobre2014oxford}. Both play a role during driving; bottom-up cues help react to signals and signs (salient by design) \cite{2014_ACP_Mccarley} and potential hazards \cite{1998_Perception_Chapman, 2003_TRF_Underwood}, and top-down strategies are important for safely steering the vehicle along the planned route. For example, at intersections, drivers display different gaze patterns depending on maneuver, presence and status of traffic signals, road layout, and actions of other road users \cite{2012_AccidentAnalysis_Werneke, 2014_CogTechWork_Werneke, 2015_TR_Lemonnier, 2019_JSR_Li}. Assistive and monitoring systems should be able to accurately capture these aspects of drivers' attention to provide adequate feedback in safety-critical situations. For example, a system which relies on a model of drivers' attention that does not implement looking both ways before entering an unsignalized intersection will not be able to check for presence of conflicting road users or notify the driver who failed to do so. 

Even though there is significant evidence for top-down effects on directing drivers' gaze, most of the existing models do not explicitly include them. Furthermore, driving datasets with attention-related data do not contain labels relevant to the task and context, thus making training models and evaluating their performance with respect to these factors difficult. To this end, here, we propose an extension of the popular DR(eye)VE dataset \cite{2018_PAMI_Palazzi}, a benchmark focusing on task- and context-relevant features, a new model combining visual and task-relevant data for predicting drivers' gaze, and experiments investigating the effects of various task and context representations.

\section{Related works}

\noindent
\textbf{Visual attention.} Research on estimating and predicting driver's gaze is rooted in the field of vision science studying bottom-up (data-driven) and top-down (task-driven) influences on human visual attention \cite{tsotsos2011computational, nobre2014oxford}. The majority of the models proposed to date perform bottom-up saliency prediction in static images and video sequences by identifying properties of rare or out-of-context visual features that attract gaze \cite{2012_TPAMI_Borji, 2021_TPAMI_Borji}. Top-down attention, on the other hand, is relatively less studied and is typically investigated within the paradigm of visual search involved in many common activities \cite{wolfe2017five}. In the early works \cite{navalpakkam2005modeling, frintrop2006vocus}, characteristics of the search target (\eg color) were used to adjust weights of bottom-up features in the output. Modern deep learning models continue to explore modulation techniques during feedforward \cite{rosenfeld2018priming, ramanishka2017top, ding2022efficient} and backward passes \cite{cao2015look, biparva2017stnet, zhang2018top} for a range of applications beyond visual search, \eg object detection, segmentation, and visuo-linguistic tasks. 

\noindent
\textbf{Driver gaze prediction.} Models for predicting drivers' gaze can likewise be subdivided by the types of attentional mechanisms they model and stimuli they use. The majority of deep learning methods achieve promising results on drivers' gaze prediction benchmarks by learning associations between images of the scene and human saliency maps in a bottom-up fashion \cite{2022_T-ITS_Kotseruba}. Image-based models use CNNs pretrained on an image classification task to encode visual information and then apply convolution and upsampling layers to generate gaze maps \cite{2019_ITSC_Ning, 2020_T-ITS_Deng}. Video-based models aim to extract spatio-temporal patterns from stacks of frames; DR(eye)VENet \cite{2018_PAMI_Palazzi}, SCAFNet \cite{2022_T-ITS_Fang}, and MAAD \cite{2021_ICCVW_Gopinath} use a 3D CNN pretrained on action classification task as an encoder, whereas BDD-ANet \cite{2018_ACCV_Xia} and ADA \cite{2022_T-ITS_Gan} extract features from each frame via a pretrained CNN and then feed them into a ConvLSTM. Many models use features in addition to visual input, such as optical flow \cite{2019_ITSC_Ning, 2018_PAMI_Palazzi, 2021_ICCVW_Gopinath}, segmentation maps \cite{2018_PAMI_Palazzi, 2020_CVPR_Pal, 2022_T-ITS_Fang}, and depth \cite{2020_CVPR_Pal}.

Top-down influences are often modeled by blending bottom-up saliency outputs with driving-task-related features, such as vanishing point \cite{2016_T-ITS_Deng, 2019_WACV_Tavakoli, 2018_T-ITS_Deng}, where drivers often maintain gaze \cite{1994_Nature_Land}, drivers' actions derived from vehicle telemetry \cite{2011_BMVC_Borji, 2014_TransSysManCybernetics_Borji}, or gaze location priors associated with certain actions \cite{2017_IV_Tawari}. A recent HammerDrive model \cite{2021_T-ITS_Amadori} recognizes maneuvers, \eg lane changes, from vehicle data and uses this information to assign weights to bottom-up saliency predictors trained on each task. MEDIRL \cite{2021_ICCV_Baee}, learns attention policy via inverse reinforcement learning, where agent state is represented by the vehicle telemetry, as well as local and global context extracted from visual input.

\noindent
\textbf{Driving datasets.} Publicly available large-scale datasets comprised of driving footage and gaze information greatly boosted research on drivers' gaze prediction. The first such dataset, 3DDS \cite{2011_BMVC_Borji}, was recorded in a video game environment. More recent ones are captured either on the road with drivers wearing head-mounted eye trackers (\eg DR(eye)VE \cite{2018_PAMI_Palazzi} and LBW \cite{2022_ECCV_Kasahara}) or use pre-recorded naturalistic footage that human subjects can view in the lab (\eg BDD-A \cite{2018_ACCV_Xia} and DADA-2000 \cite{2019_ITSC_Fang}). However, task-relevant information that can be effectively used for training and evaluation is usually not available or incomplete. For example, only parts of DR(eye)VE dataset with unusual gaze patterns are annotated as ``acting'' without specification of the action being performed by the driver. Similarly, in BDD-A, ground truth deviating from the mean is associated with drivers' actions or sudden events without explicit class labels. The authors of \cite{2017_IV_Tawari} and \cite{2021_ICCV_Baee} evaluate models on different types of scenarios from DR(eye)VE and BDD-A dataset, but do not make the annotations available. 

\noindent
\textbf{Contributions:} 1) we correct the data processing pipeline in DR(eye)VE to reduce noise in the recorded gaze data; 2) we address the lack of consistent task-relevant annotations by labeling drivers' actions and relevant context elements; 3) using the new annotations and ground truth, we benchmark to analyze baseline and state-of-the-art (SOTA) models to analyze their performance on the entire dataset and with respect to task and context factors; and 4) we propose SCOUT, a novel task- and context-aware model, that demonstrates advantages of task-relevant information for modulating bottom-up driver gaze prediction.

\section{Extending DR(eye)VE}
We chose DR(eye)VE \cite{2018_PAMI_Palazzi} since it is recorded in real road conditions \textit{and} provides all raw data, including eye tracker output and ego-vehicle telemetry. This combination of ecological validity and rich data makes it the most suitable starting point among datasets listed in the previous section.

\subsection{Original ground truth}

DR(eye)VE consists of $~6$ hours ($0.55$M frames) of driving footage and gaze data recorded on-road in an actual vehicle. Eight participating drivers wore eye-tracking glasses (ETG) which allowed free head movement. Due to the insufficient quality of the eye tracker world-facing camera, its limited view, and drivers' head movements, an additional wide-angle Garmin camera (GAR) was installed on top of the vehicle's roof. Ground truth saliency maps were created following a two-stage procedure: 1) drivers' gaze coordinates were mapped from driver's view (ETG) to the rooftop camera view (GAR) via a geometric transformation; 2) since each recording was associated with a single driver's gaze, only a handful of data points were recorded per frame. Thus, to reduce sparsity and individual drivers' biases, data points were aggregated over a sliding window of 25 frames (1s). 

\subsection{New ground truth}

As we examined the raw eye-tracker output and ground truth data in DR(eye)VE, several irregularities were identified: 1) temporal misalignment of the world-view eye-tracking glasses camera (ETG) and rooftop camera (GAR) video streams; 2) incorrect inclusion of saccades and blinks in the ground truth; 3) incorrect inclusion of drivers' gaze towards the vehicle interior and outside the rooftop camera view; 4) noisy homography transformations for mapping gaze from ETG to GAR and aggregating gaze across GAR frames; and 5) artifacts caused by the Gaussian response function used to compute the original ground truth. Below, we discuss how we addressed each of these shortcomings.

\begin{figure}[t!]
\centering
\includegraphics[width=\columnwidth]{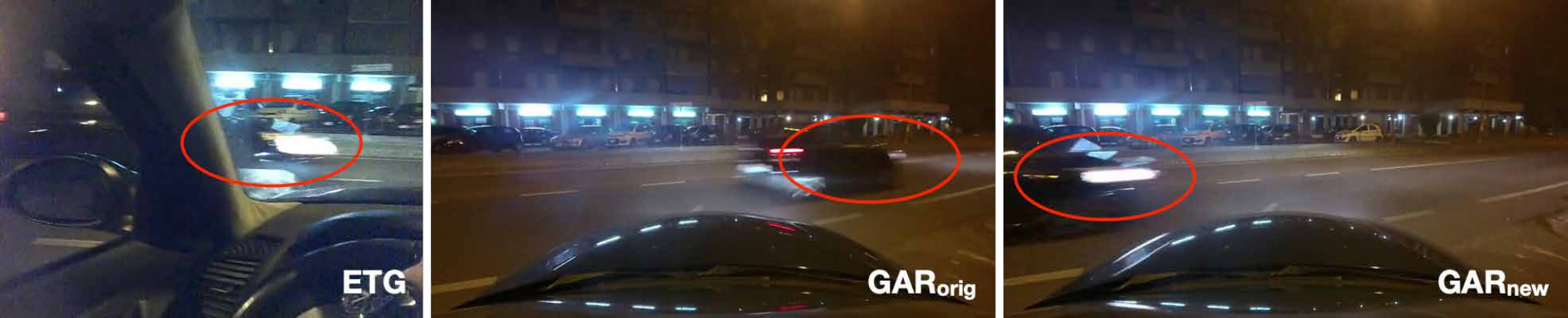}
\caption{Example of misalignment by 9 frames. The car in the driver's view (ETG) is to the left of the ego-vehicle but is to the right in the rooftop camera view (GAR\textsubscript{orig}). Our recomputed GAR\textsubscript{new} shows a correct alignment.}
\label{fig:alignment}
\vspace{-1em}
\end{figure}

\begin{figure*}[thbp!]
\centering
\includegraphics[width=\textwidth]{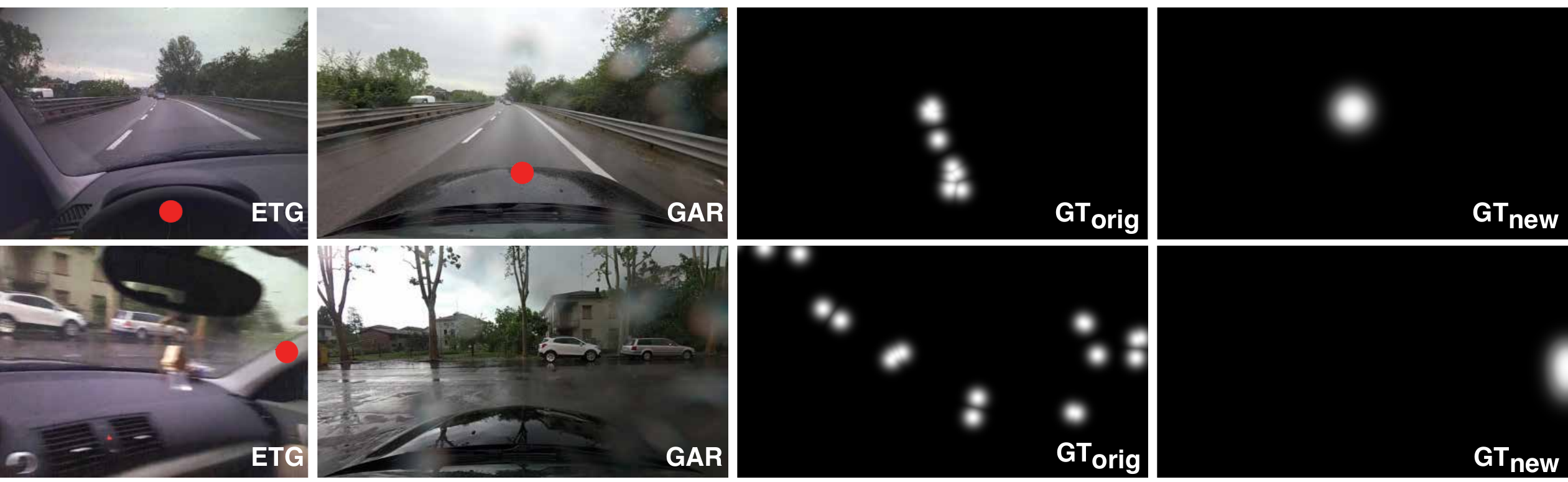}
\caption{Typical sources of noise in the original DR(eye)VE ground truth and effects of the proposed corrections. ETG and GAR refer to driver's and rooftop camera views, respectively. Fixations (red circles) are shown for time $t$. To form the ground truth (GT), fixations are aggregated over 1s (or $t\pm12$). To compensate for motion in the frame, the original ground truth (GT$_{\mathrm{orig}}$) uses homography transformation and the new one (GT$_{\mathrm{new}}$) relies on optical flow, resulting in less noisy saliency maps. Top row: The vertical ``trail'' of points in GT$_{\mathrm{orig}}$ corresponds to a saccade towards the speedometer and is absent in GT$_{\mathrm{new}}$ after removal of saccades and in-vehicle fixations. Bottom row: Incorrect mapping of driver's gaze outside the GAR view results in a random pattern in GT$_{\mathrm{orig}}$. In GT$_{\mathrm{new}}$, such out-of-frame fixations are pushed towards the edge of the frame to preserve the direction and elevation of the driver's gaze. Additional examples are provided in Appendix \protect \ref{suppl:gt}.}
\label{fig:old_new_gt}
\vspace{-1em}
\end{figure*}

\noindent
\textbf{Video re-alignment.} Due to different frame rates of the cameras, each ETG and GAR video contains 9000 and 7500 frames, respectively. The original mapping between ETG and GAR frames is consistent with linear interpolation. % run script create/ground_truth/check_alignment.m
However, upon closer inspection, we found that videos were not aligned temporally (Figure \ref{fig:alignment}); often, the GAR camera continued recording up to 15--20 frames (0.5-0.8s) after the ETG camera stopped. In some videos, the cameras also did not start recording at the same time. Since videos did not contain timestamps, we manually aligned them using events visible in both camera views, such as windshield viper movement, traffic light changes, \etc \ Most videos were misaligned by $7\pm5$ frames on average, $40\%$ of the videos by up to 15 frames (0.5s) and $10\%$ by over 25 frames (1s). %The effects of asynchrony are naturally more noticeable in busy urban scenes with other vehicles passing the ego-vehicle, but are also apparent in the relatively ``empty'' rural and night sequences too.

%\begin{figure}[t!]
%\centering
%\includegraphics[width=0.6\columnwidth]{gaze_types_sunburst_v1.pdf}
%\caption{Composition of eye-tracking data in DR(eye)VE. Inner circle: \% of data points labeled as fixations, blinks, and saccades. Outer circle: spatial distribution of fixations.}
%\label{fig:dreyeve_gaze_event_types}
%\vspace{-1em}
%\end{figure}

%\begin{figure}[t!]
%\centering
%\includegraphics[width=\columnwidth]{intersection.pdf}
%\caption{At intersections, the driver may look outside the  in the original ground truth, points representing saccades in the past frames and out-of-bounds fixations are still projected onto the scene in a random pattern. In the new ground truth, to preserve some gaze information, we push out-of-bounds fixations towards the edge of the frame to indicate the approximate direction of where the driver is attending.}
%\label{fig:out_of_bounds_gaze}
%\vspace{-1em}
%\end{figure}

\noindent
\textbf{Removal of saccades, blinks, and in-vehicle fixations, remedying out-of-bounds fixations.}  Eye-tracking glasses used for data collection provide labels for fixations, saccades, blinks, and recording errors. Although it is common to filter out saccades and blinks from ground truth, saliency maps provided with DR(eye)VE contain both. Since during blinks and saccades vision is largely suppressed  \cite{ross2001changes}, inclusion of such data effectively biases the dataset by implying that the driver perceived something at those locations. Saccades also produce ``trails'' in the ground truth maps due to temporal aggregation (see Figure \ref{fig:old_new_gt}, top row). Therefore, all non-fixations are excluded from the new ground truth.

In DR(eye)VE, $95\%$ of fixations fall on the windshield and only a small fraction is directed at the vehicle interior or towards the side windows: 

\begin{itemize}[topsep=0pt,itemsep=0pt,parsep=0pt,partopsep=0pt,leftmargin=0.5em]

\item \textit{In-vehicle gaze.} Nearly 3\% of all fixations are directed towards the speedometer, mirrors, and passengers in the vehicle. %(with significant individual variation across drivers).
When projected onto the scene, these gaze points will incorrectly reflect what the driver was looking at. 
%Originally, the authors of the dataset provided a horizontal threshold for each ETG video which was used to eliminate in-vehicle fixations, however, it only removes some of the fixations towards the speedometer and dashboard. On the other hand, this method works poorly when 1) driver's gaze shifts vertically with respect to the dashboard as a result of vehicle movement or driver's head movement, 2) driver looks at the side, rear-view mirrors, or towards the passenger, areas that are above the dashboard and are not always visible in the frame. 
We manually annotated fixations towards vehicle interior and removed those irrelevant to the driving scene (\eg speedometer, passengers) from the ground truth.

\item \textit{Errors.} Fixations that were far out of bounds in the ETG camera view (1\% of the total) were removed as they could not be reliably mapped to GAR view. 

\item \textit{Gaze outside the camera viewpoint.} When the driver turns their head, \eg to look through side windows or over the shoulder, their gaze often falls outside the rooftop camera (GAR) view (see Figure \ref{fig:old_new_gt}, bottom row). In the original dataset, these fixations were erroneously mapped onto the scene (causing random patterns in ground truth) or manually removed (resulting in blank ground truth) but not marked in annotations. We manually labeled all such fixations. Since they are present in virtually all safety-critical scenarios (\eg intersections), in the new ground truth, we push out-of-frame fixations towards the edge of the image to preserve the direction and elevation of driver's gaze.
%[example of ETG frame, GAR frame, old and new gt]

\end{itemize}

\noindent
\textbf{Re-computing and manual correction of projected fixations.} Video re-alignment changed $\approx90\%$ of the correspondences between ETG and GAR frames, therefore we could no longer use the pre-computed homographies supplied with the dataset to map gaze from ETG to GAR view. We used the VLFeat library \cite{vedaldi2010vlfeat} with the same parameters and thresholds as in the original implementation to compute homography transformations for the new alignment, inspected the results, and manually corrected all outliers. %We then used the following procedure to estimate the error: for pairs of randomly selected ETG-GAR frames with fixations, we computed a homography transformation 10 times and applied the results to the fixations. We then computed the Euclidean distance from new fixations to manual ground truth. The results are plotted in [DIAGRAM].

%In addition, we measured the quality of the original homographies between ETG and Garmin cameras. Analysis of the homography matrices provided by the authors revealed that $18\%$ of them result in anomalous rescaling ($>3\times$), $17\%$ are missing, $10\%$ are singular, and another $2\%$ do not preserve orientation (\ie result in mirror images). For the night videos, however, the rates were worse, on average only $35\%$ homographies were acceptable, with multiple videos only having $10-15\%$ usable homographies. Due to central bias of fixations, even poor homographies generally preserved them, however, this also exacerbated errors on off-center fixations.

\noindent
\textbf{Data re-aggregation.} Although use of homography is justified in case of ETG-GAR transformation, it is less so for aggregating GAR frames across a 1s temporal window. %We use a similar procedure as described in previous subsection to measure variations caused by GAR-GAR transformation. Since computing ground truth was prohibitive we instead used a centroid of fixations to compute deviation. The results indicate that this procedure results in $10-15\%$ of outliers. As expected, the number of errors and their magnitude grow as temporal distance from the keyframe increases. [DIAGRAM]
Since the main purpose of this transformation is to compensate for motion in the scene, we instead apply a technique from \cite{2013_TVCG_Kurzhals} that relies on optical flow. We use RAFT \cite{teed2020raft} to compute optical flow and then use its magnitude and direction to trace locations of fixations from frames $t-12, ..., t-1$ and $t+1, ..., t+12$ to the key frame $t$.

\noindent
\textbf{Generating heatmaps.} The original ground truth is computed as a max over Gaussian response function for each data point in the temporal window. Here, we set the fixated locations to $1$ and apply an isotropic Gaussian filter with the size of $40$ px, roughly equal to human fovea radius following \cite{mital2011clustering, jiang2015salicon, xu2014predicting, nguyen2018your}. The final result is shown in Figure \ref{fig:old_new_gt}.

\begin{table}[t!]
\centering
\caption{Counts, durations (in frames), and \% of frames containing lateral and longitudinal actions, and intersections}
\resizebox{0.85\columnwidth}{!}{%
\begin{tabular}{@{}clcccc@{}}
\multicolumn{2}{c}{Actions/context} & Count & Mean & Std & \% frames \\ \midrule
\multicolumn{1}{c|}{\multirow{5}{*}{\rotatebox{90}{\begin{tabular}[c]{@{}c@{}}Lateral \\ actions\end{tabular}}}} & Drive straight & 499 & 1034.43 & 1308.38 & 93.0 \\
\multicolumn{1}{c|}{} & Turn right & 142 & 68.32 & 33.24 & 1.8 \\
\multicolumn{1}{c|}{} & Lane change right & 117 & 80.86 & 39.61 & 1.7 \\
\multicolumn{1}{c|}{} & Lane change left & 119 & 79.00 & 38.12 & 1.7 \\ 
\multicolumn{1}{c|}{} & Turn left & 93 & 94.44 & 25.51 & 1.6 \\ \midrule
\multicolumn{1}{c|}{\multirow{4}{*}{\rotatebox{90}{\begin{tabular}[c]{@{}c@{}}Long.\\ actions\end{tabular}}}} & Maintain & 793 & 441.84 & 867.92 & 63.1 \\
\multicolumn{1}{c|}{} & Accelerate & 486 & 190.88 & 112.52 & 16.7 \\
\multicolumn{1}{c|}{} & Decelerate & 483 & 156.93 & 75.73 & 13.7 \\
\multicolumn{1}{c|}{} & Stopped & 108 & 333.85 & 394.5 & 6.5 \\ \midrule
\multicolumn{1}{c|}{\multirow{4}{*}{\rotatebox{90}{\begin{tabular}[c]{@{}c@{}}Inters.\\type\end{tabular}}}} & Unsignalized & 541 & 47.74 & 47.42 & 3.65 \\ 
\multicolumn{1}{c|}{} & Merge & 159 & 33.28 & 20.18 & 0.95 \\
\multicolumn{1}{c|}{} & Signalized & 106 & 48.74 & 85.94 & 0.93 \\
\multicolumn{1}{c|}{} & Roundabout & 58 & 63.83 & 19.55 & 0.67 \\ \bottomrule
\end{tabular}%
}
\label{tab:lat_lon_action_stats}
\vspace{-1.5em}
\end{table}

\subsection{Extended annotations}
\label{sec:task_context_annotations}
To analyze the effects of tasks and context on gaze, we extended the annotations. Here, \textit{task} is a series of maneuvers (actions) that the drivers perform on the route to their destination. \textit{Context} is defined as the intersections along the route and drivers' priority during the maneuver, \ie whether they have a right-of-way or must yield to other road users.

\noindent
\textbf{Task annotations.} Drivers' actions were divided into lateral and longitudinal and annotated per-frame. For lateral actions, we identified and manually labeled \textit{left/right turns} and \textit{lane changes}, as well as \textit{U-turns}. The rest were labeled as \textit{driving straight}. To produce longitudinal action labels, we first interpolated and smoothed vehicle speed with an average filter, and computed acceleration from these values. \textit{Stopped} labels were assigned to segments where speed did not exceed $1$ km/h, threshold for \textit{acceleration} and \textit{deceleration} labels was set to $\pm0.4$ m/s$^2$. Table \ref{tab:lat_lon_action_stats} shows statistics of the action labels in the dataset. Less than $7\%$ of all frames contain lateral actions, whereas longitudinal actions take place in $27\%$ of the frames, and only $4\%$ of the frames contain both longitudinal and lateral action.

\noindent
\textbf{Context annotations.} Context annotations include \textit{intersection locations} and \textit{types}, as well as drivers' \textit{priority} while passing the intersection. The following intersection types are annotated: \textit{roundabouts}, \textit{highway on-ramps}, \textit{signalized} (controlled by traffic lights), and \textit{unsignalized} (sign-controlled or uncontrolled) intersections. For intersections, where the driver yielded to opposing traffic, we count frames from the first time that the driver fixated on the intersecting road. When the driver has a right-of-way, we consider preceding 25 frames, due to the approx. $1s$ gaze-action delay \cite{2017_FPsych_Lappi}. Table \ref{tab:lat_lon_action_stats} shows the counts and duration of video segments that contain intersections, accounting for $\approx7\%$ of all frames. Action and context categories are balanced across training and test sets.

\vspace{-0.5em}
\section{Benchmark results}
\label{sec:benchmark}

\begin{table}[t!]
\centering
\caption{Results on the original and new DR(eye)VE ground truth. Arrows  indicate that larger ($\uparrow$) and smaller ($\downarrow$) values are better. Best and second-best values are shown as \textbf{bold} and \underline{underlined}.}
\vspace{-0.5em}
\includegraphics[width=\columnwidth]{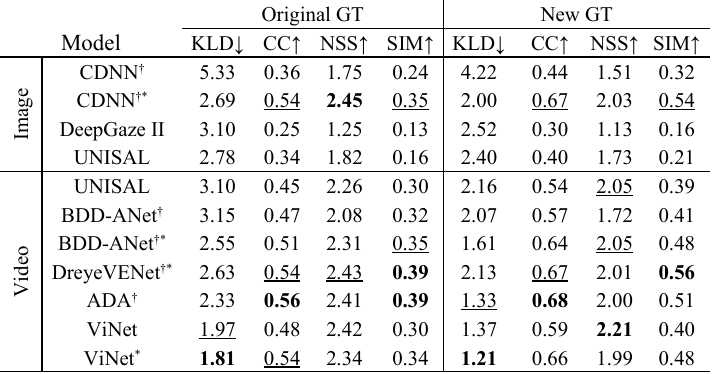}
\vspace{-1.5em}
\begin{flushleft}
\begin{footnotesize} *---model is  trained on data; \textdagger---model for drivers' gaze prediction.
\end{footnotesize}
\end{flushleft}
\vspace{-2em}
\label{tab:benchmark_all}
\end{table}

In this section, we evaluate a number of models including SOTA generic bottom-up saliency algorithms (DeepGaze II \cite{kummerer2016deepgaze}, UNISAL \cite{2020_ECCV_Droste}, ViNet \cite{2021_IROS_Jain}) and drivers' gaze prediction algorithms (CDNN \cite{2016_T-ITS_Deng}, DReyeVENet \cite{2018_PAMI_Palazzi}, BDD-ANet \cite{2018_ACCV_Xia}, and ADA \cite{2022_T-ITS_Gan}) that operate on image and video data (UNISAL works on both). We report the evaluation results using common saliency metrics: Normalized Scanpath Saliency (NSS), histogram similarity (SIM), Pearson's correlation coefficient (CC), and Kullback--Leibler divergence\footnote{We compute KLD with code in \cite{fahimi2021metrics}, which is more numerically stable than implementation in \cite{2018_PAMI_Palazzi} used to produce results reported in the literature. Note that only the absolute KLD values change, the relative ranking of the models is preserved. For more details see Appendix \ref{suppl:KLD}.} (KLD) \cite{2018_PAMI_Bylinskii}. We train and evaluate the models on original and new ground truth. Blank ground truth frames and frames that contain U-turns are excluded from evaluation\footnote{Blank gt maps result from gaze data missing due to blinks\slash saccades\slash errors and may cause issues with some metrics. U-turns are removed because there are only 14 instances, all in dead ends with no traffic, and most gaze information is lost since the driver looks at the mirrors or over the shoulder.}.

We first compare the model performance on the entire dataset to examine the effect of cleaning up the data. Next, using the corrected ground truth and new task and context annotations, we analyze performance of the models with respect to various aspects of task and context, namely, driver's actions, intersections types, and ego-vehicle priority when passing through them. Due to space limitations we report the results only on KLD metric. Evaluation on other metrics (CC, NSS, and SIM) on actions and context subsets reveals similar patterns as KLD shown. Due to limited space, these results will be released with the code for the paper. 

\noindent
\textbf{Overall performance.} Results presented in Table \ref{tab:benchmark_all} lead to two observations. First, performance of all models improves significantly on all metrics when trained or evaluated on the new ground truth data, which can be attributed mainly to the removal of a large number of blinks and saccades as well as to the reduction of artifacts introduced by the original fixation aggregation procedure (see Figure \ref{fig:old_new_gt}). Second, generic saliency models (particularly image-based ones) do not generalize well to driving data, but when trained, can perform on par or better than driving-specific gaze prediction models. For example, ViNet outperforms driving-specific models on some metrics despite not using any additional information (\eg semantic segmentation and optical flow as in DReyeVENet) or weighted sampling (as in BDD-ANet). 

%However, the number of nonzero ground truth frames is different in the original and new ground truth due to video re-alignment, removal of blinks and saccades, and re-aggregation of fixations (see Section). Therefore, we report the results on two sets: \textit{old+new} -- valid (non-empty) ground truth frames in both orig and new data and \textit{new} -- valid ground truth in new data to establish a baseline for the next section. 

\begin{table}[t!]
\centering
\caption{Results on the \textit{new} DR(eye)VE ground truth for different types of actions: None---maintain speed/lane, Acc---accelerate, Dec---decelerate, Lat---lateral actions only, Lat/lon---simultaneous lateral and longitudinal actions, Stop---ego-vehicle is stopped. Green and red colors indicate the best and worst values. Best and second-best values are \textbf{bold} and \underline{underlined}.}
\vspace{-1em}
\includegraphics[width=0.7\columnwidth]{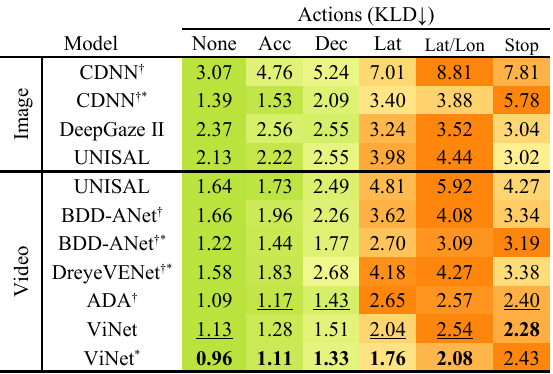}
\vspace{-1em}
\begin{flushleft}
\begin{footnotesize} *---model is  trained on data; \textdagger---model for drivers' gaze prediction. Results on other metrics are given in Table \ref{tab:benchmark_extra}.
\end{footnotesize}
\end{flushleft}
\vspace{-1em}
\label{tab:benchmark_task}
\end{table}

\begin{table}[t!]
\centering
\caption{Results on the \textit{new} DR(eye)VE ground truth for different types of intersections and priorities (RoW---right-of-way). Green and red colors indicate the best and worst values. Best and second-best values are \textbf{bold} and \underline{underlined}.}
\vspace{-1em}
\includegraphics[width=0.9\columnwidth]{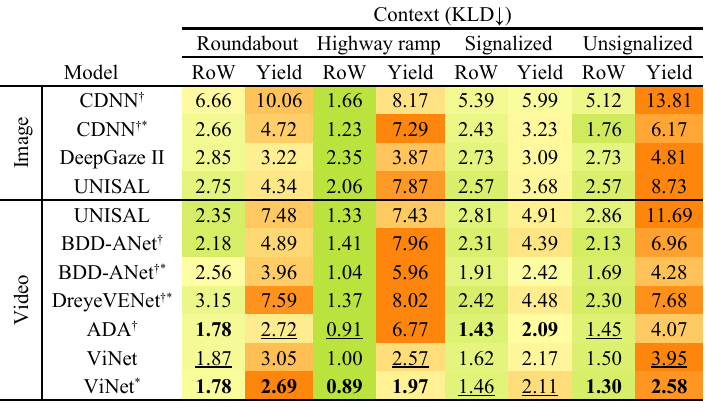}
\vspace{-1em}
\begin{flushleft}
\begin{footnotesize} *---model is  trained on data; \textdagger---model for drivers' gaze prediction. Results on other metrics are given in Table \ref{tab:benchmark_extra}.
\end{footnotesize}
\end{flushleft}
\vspace{-2.5em}
\label{tab:benchmark_context}
\end{table}

\noindent
\textbf{Results on different types of actions.} In Table \ref{tab:benchmark_task}, the results are aggregated over different types of actions, as well as times when the ego-vehicle is not moving or is maintaining lane\slash speed. Naturally, predicting \textit{stop} sequences is more challenging, since drivers' gaze tends to wander around the scene. In sequences where drivers maintain speed and lane (\textit{None}), they predominantly look at the vanishing point or vehicle ahead. As a result, the gaze is very centered and performance of all models is high across all metrics. Longitudinal actions (\textit{Lon}) pose less challenge than lateral actions since they still require the driver to primarily attend to the objects located in front, such as the lead vehicle and traffic signals or signs. Lateral actions (\textit{Lat}) often involve checking mirrors and scanning across the scene, therefore are more difficult to predict. Finally, both lateral and longitudinal maneuvers (\textit{Lat+Lon}) are performed when crossing intersections or interacting with other road users. In such cases, gaze patterns are more complex because drivers need to look out for potential conflicting vehicles and pedestrians. If areas of interest are not visible to the scene camera due to its limited field of view, models lack visual cues to associate with the recorded drivers' gaze (see Figure \ref{fig:old_new_gt}). Since such scenarios are both highly variable and the least represented in the training set, thus performance of all models on this subset is reduced across all metrics.

\noindent
\textbf{Results on different types of context.} Table \ref{tab:benchmark_context} shows results aggregated by four intersection types (roundabouts, highway ramps, signalized, unsignalized) and two priority options (right-of-way and yield). Here, we observe that sequences where ego-vehicle must yield to other road participants leads to a dramatic drop in performance for all models. Yielding scenarios are generally more complex that right-of-way ones, since the driver must check different areas of the road (\eg look to the left and right for incoming traffic), often in advance, and irrespective of presence of other road users. Although in DR(eye)VE traffic density is light, there is still a lot of variability in terms of possible scenarios in each context and a small number of samples in the training data, the bottom-up learning strategy used in all tested models struggles to capture such gaze patterns. 

Not all intersection types are equally challenging. Roundabouts and unsignalized intersections are more difficult since they always require the driver to check the surroundings and look out for other road users. Whereas when passing signalized intersections or merging on a highway, gaze patterns are simpler since drivers tend to observe the same areas, \eg looking towards the lane immediately to the left when entering highway on-ramp or checking the traffic approaching from the opposite direction when making a left turn.

\section{SCOUT: Ta\underline{s}k- and \underline{c}ontext-m\underline{o}d\underline{u}lated a\underline{t}tention}

%exp_config_2_2_2023_08_27_20_05_15 tested weighted loss and weighted sampling (with raw KLdiv values)

%SCOUT	taSk Context mOdUlated aTtention
%AUTO	Attention modUlated Task cOntext
%ACTUATE	Action ContexT modUlated ATtEntion
%TOOT	acTion cOntext mOdulated aTtention
%NOMAD	actioN cOntext Modulated gAze preDiction
%CAUTION	Context Action modUlated aTtentION

Next, we explore integration of the new task and context annotations for modulating gaze prediction. The problem is formulated as follows: given a set of video frames and a set of labels representing task and context information, predict a human gaze map for the last frame of the set. 

\subsection{Model architecture}

\begin{figure}[t!]
\centering
\includegraphics[width=\columnwidth]{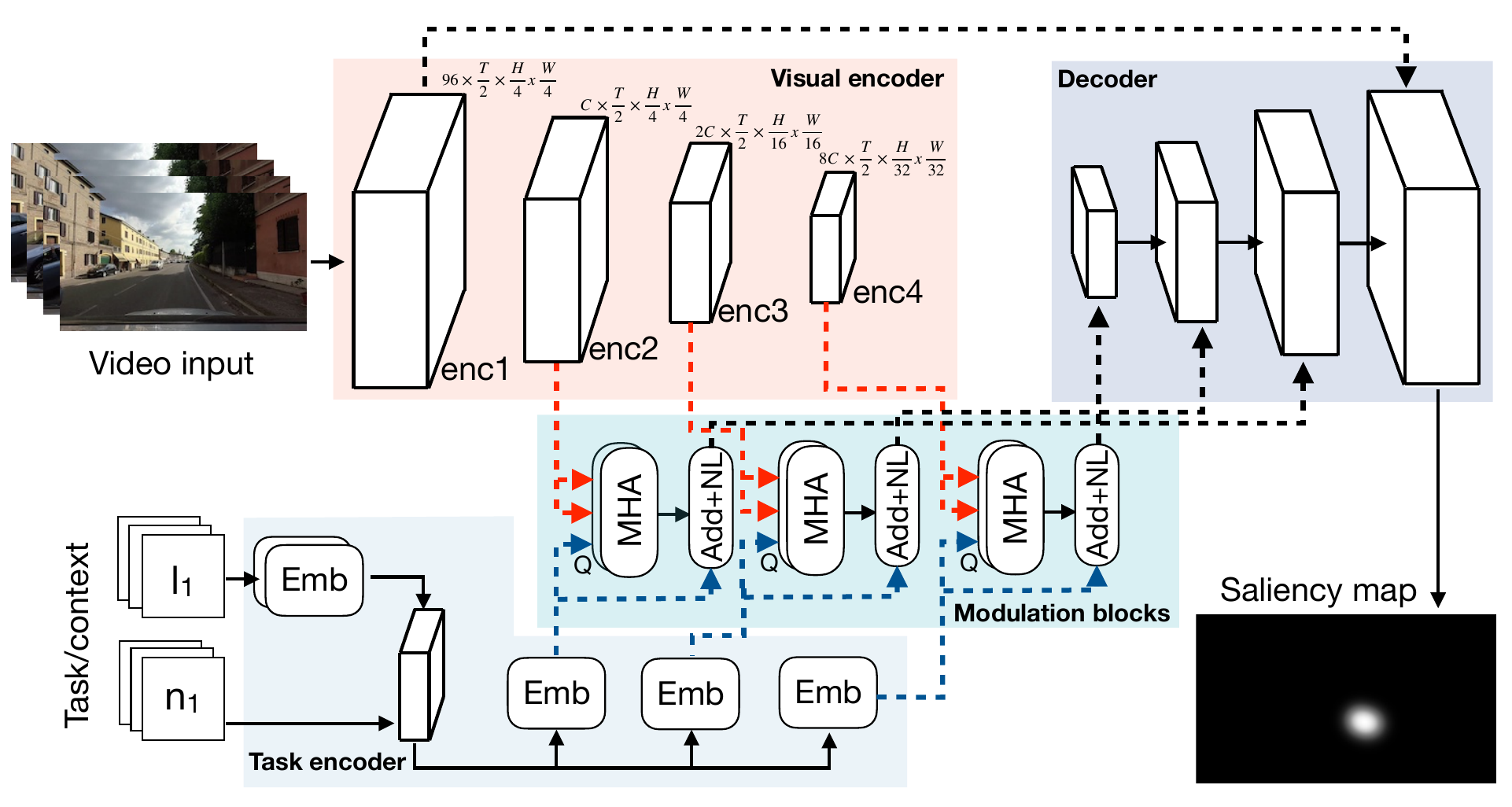}
\caption{Architecture of SCOUT model: two encoders for visual and task/context data, multi-head attention (MHA) modulation blocks, and a decoder. Text labels $l_1, l_2, \dots l_n$ and numeric features $n_1, n_2, \dots, n_m$ represent task and context information. Dashed arrows indicate possible connections between modules depending on how many modulation blocks are used for fusing visual and task/context features.}
\label{fig:architecture}
\vspace{-1.5em}
\end{figure}

\begin{table*}[t!]
\centering
\caption{Performance of SCOUT model on the \textit{new} DR(eye)VE ground truth without task and with task and context information. For reference, we show two models, ADA and ViNet, whose results were the best according to the benchmark in the previous section. $\uparrow$ and $\downarrow$ indicate that larger and smaller values are better. Best and second-best values are shown as \textbf{bold} and \underline{underlined}, respectively. Abbreviations:None---maintain speed/lane, Acc---accelerate, Dec---decelerate, Lat---lateral actions only, Lat/lon---simultaneous lateral and longitudinal actions, Stop---ego-vehicle is stopped, GC --- global context, LC -- local context, CA -- current action. Numbers in square brackets [] specify encoder blocks at which task information was added. Results on other metrics are given in Table \ref{tab:SCOUT_extra}.}
\vspace{-0.5em}
\setlength{\tabcolsep}{0.25em}
\resizebox{\textwidth}{!}{%
\begin{tabular}{cccccccccccccccccccccc}
\multicolumn{1}{l}{}     & \multicolumn{1}{l}{}    & \multicolumn{1}{l}{} & \multicolumn{1}{l}{} & \multicolumn{1}{l}{} & \multicolumn{1}{l}{} & \multicolumn{1}{l}{} & \multicolumn{1}{l}{} & \multicolumn{1}{l}{} & \multicolumn{1}{l}{} & \multicolumn{1}{l}{} & \multicolumn{1}{l}{} & \multicolumn{1}{l}{} & \multicolumn{8}{c}{Context (KLD$\downarrow$)}       \\ \cline{15-22} 
\multicolumn{2}{c}{}   & \multicolumn{4}{c}{All}      &   & \multicolumn{6}{c}{Actions (KLD$\downarrow$)}     &   & \multicolumn{2}{c}{Roundabout} & \multicolumn{2}{c}{Highway ramp} & \multicolumn{2}{c}{Signalized} & \multicolumn{2}{c}{Unsignalized} \\ \cline{3-6} \cline{8-13} \cline{15-22} 
      & Model     & KLD$\downarrow$    & CC$\uparrow$     & NSS$\uparrow$    & SIM$\uparrow$    &   & None & Acc & Dec & Lat & Lat/Lon & Stop &   & RoW     & Yield      & RoW   & Yield     & RoW  & Yield    & RoW   & Yield     \\ \cline{1-6} \cline{8-13} \cline{15-22} 
      & ADA\textdagger  & 1.33 & 0.68 & 2.00 & 0.51 &  & 1.09 & 1.17 & 1.43 & 2.65 & 2.57 & 2.40 &  & 1.78 & 2.72 & 0.91 & 6.77 & 1.43 & 2.09 & 1.45 & 4.07 \\
      & ViNet*     & 1.21 & 0.66 & 1.99 & 0.48 &  & 0.96 & 1.11 & 1.33 & 1.76 & 2.08 & 2.43 &  & 1.78 & 2.69 & 0.89 & \uline{1.97} & 1.46 & 2.11 & 1.30 & 2.58 \\ \cline{1-6} \cline{8-13} \cline{15-22} 
\multicolumn{1}{c|}{\multirow{7}{*}{\rotatebox{90}{SCOUT}}} & w/o task & 1.08 & 0.63 & 3.89 & 0.52 &  & 0.81 & 0.99 & 1.25 & 1.62 & 1.99 & 2.43 &  & \textbf{1.26} & 2.89 & 0.74 & 2.24 & 1.15 & 2.46 & 1.10 & 3.11 \\
\multicolumn{1}{c|}{}    & w/o task+weight. loss  & 1.01 & 0.65 & 4.04 & 0.51 &  & 0.77 & 0.93 & 1.15 & 1.55 & 1.88 & \uline{2.05} &  & 1.41 & 2.60 & 0.72 & 2.26 & 1.09 & 1.88 & 1.02 & 2.74 \\
\multicolumn{1}{c|}{}    & w/o task+weight. sampl.  & 1.01 & 0.65 & 4.02 & 0.52 &  & 0.76 & 0.94 & 1.14 & 1.55 & 1.86 & 2.24 &  & 1.51 & 2.61 & 0.66 & 2.12 & 1.16 & 1.77 & 1.01 & 2.62   \\ \cline{2-6} \cline{8-13} \cline{15-22} 
\multicolumn{1}{c|}{}    & w/task CA {[}3{]}   & 0.97 & 0.66 & 4.06 & 0.52 &  & \uline{0.73} & 0.91 & 1.10 & 1.54 & 1.83 & \uline{2.05} &  & \uline{1.29} & 2.43 & \uline{0.64} & 2.29 & 1.08 & 1.90 & 1.04 & 2.42 \\
\multicolumn{1}{c|}{}    & w/task GC+LC+CA {[}2,3,4{]} & 0.95 & 0.66 & 4.09 & \uline{0.54} &  & \uline{0.73} & \uline{0.88} & \uline{1.06} & 1.45 & 1.75 & 1.95 &  & 1.58 & \uline{1.99} & \uline{0.64} & 2.21 & 1.03 & 2.12 & \uline{0.93} & \uline{2.13} \\
\multicolumn{1}{c|}{}    & w/task GC+LC+CA {[}3{]}     & \uline{0.94} & \textbf{0.67} & \uline{4.13} & \uline{0.54} &  & \textbf{0.70} & 0.89 & \textbf{1.05} & \uline{1.44} & \uline{1.67} & 2.09 &  & 1.52 & \textbf{1.90} & \textbf{0.57} & 2.07 & \uline{1.01} & \textbf{1.50} & 0.97 & 2.14 \\
\multicolumn{1}{c|}{}    & w/task LC {[}2{]}     & \textbf{0.92} & \textbf{0.67} & \textbf{4.17} & \textbf{0.55} &  & \textbf{0.70} & \textbf{0.86} & \textbf{1.05} & \textbf{1.40} & \textbf{1.62} & \textbf{2.03} &  & 1.43 & 2.03 & 0.67 & \textbf{1.95} & \textbf{0.99} & \uline{1.60} & \textbf{0.89} & \textbf{2.09}  \\ \cline{1-6} \cline{8-13} \cline{15-22} 
\end{tabular}%
}

\label{tab:SCOUT_results}

\begin{flushleft}
\begin{footnotesize} *---model is  trained on the data; \textdagger---model for drivers' gaze prediction. 
\end{footnotesize}
\end{flushleft}
\vspace{-2em}
\end{table*}

The architecture of SCOUT comprises three modules: an encoder for visual and task/context features, modulation blocks, and a decoder (as shown in Figure \ref{fig:architecture}).

\noindent
\textbf{Encoder.} To encode video data, we use Video Swin Transformer (VST) \cite{2022_CVPR_Liu} as a backbone for spatio-temporal feature extraction. Compared to C3D \cite{2015_ICCV_Tran} and S3D \cite{xie2017rethinking} backbones used in other saliency models \cite{2018_PAMI_Palazzi,2021_IROS_Jain}, VST is lightweight and performs better on action recognition benchmarks, which is useful for transfer learning. VST takes a set of 16 frames as input and outputs features from each of its four blocks. 

Task and context input consists of per-sample text labels and per-frame numeric values (indicated as $l_1, \dots l_n$ and $n_1, \dots, n_m$ in Figure \ref{fig:architecture}), representing \textit{global context} (weather, time of day, and location labels provided with DR(eye)VE), \textit{local context} (next lateral action label, distance to next intersection (m) derived from GPS, and priority label), and \textit{current action} (speed in km/h, acceleration in m/s\textsuperscript{2}, and lateral action label). See Appendix \ref{suppl:task_context} for more details on task and context representation. We use a linear embedding layer to encode text labels and downsample the numeric values along the temporal dimension to match the temporal dimension of VST outputs. We then stack the encoded features, apply an affine transformation to match the feature dimension $C$ of the corresponding encoder block and replicate the features across that block's spatial dimensions. 

\noindent
\textbf{Modulation blocks.} We modulate spatio-temporal visual features with task and context information via cross-attention using multi-head attention (MHA) \cite{2017_NeurIPS_Vaswani}, following the success of this technique for modulating attention with visual \cite{ding2022efficient} and multi-modal features \cite{rasouli2023pedformer}. Encoded task features are passed to the MHA as queries, and visual features are used as keys and values. We then add the initial task features to the MHA output and apply layer normalization (LN). 

\noindent
\textbf{Decoder.} The decoder gradually fuses information from encoder and modulation blocks, while increasing their spatial dimension and reducing the channel and temporal dimensions. Each of the decoder blocks consists of one upsampling layer, ReLU layer, and 3D convolution layer. The output of each encoder or modulation block is first upsampled and concatenated with the preceding block. The result is fed through ReLU layer and then into the 3D convolutional layer.

\subsection{Implementation details} 

We use the Swin-S Video Swin Transformer model \cite{2022_CVPR_Liu} pre-trained on the Kinetics-400 dataset \cite{2017_CVPR_Carreira} as encoder backbone and continue to update its weights during training. The input to the network is a set of 16 consecutive frames ($\approx 0.5$s observation) resized to $224\times224\times3$. MHA blocks used for modulating visual data with task and context information each have 2 heads and embedding sizes set to match the $C$ dimension of the corresponding encoder blocks. We use the first 34 videos from DR(eye)VE dataset for training, videos 35--37 for validation, and the remaining videos for testing. Training samples are generated by splitting the videos into 16-frame segments with 8 frame overlap. Sequences with blank ground truth and U-turns are excluded from training, validation, and testing, as explained earlier in Section \ref{sec:benchmark}. The model is trained on a single NVIDIA Titan 1080Ti GPU for 20 epochs with the KLD loss, Adam optimizer \cite{2014_arXiv_2014}, early stopping based on validation loss, a constant learning rate of $1e-4$, and a batch size of $4$.

\subsection{Results} 
Table \ref{tab:SCOUT_results} shows the evaluation results of the SCOUT model variants with and without task information on the entire DR(eye)VE dataset, as well as subsets of it corresponding to different types of actions and context as defined in Section \ref{sec:task_context_annotations}. ADA and ViNet, models with the strongest results on action/context in Section \ref{sec:benchmark}, are shown for reference. 

\noindent
\textbf{Performance without task information.} We begin by establishing the baseline performance of the proposed model without task information. SCOUT without task improves SOTA results on all metrics, except CC, by a large margin. In particular, it improves NSS (highly sensitive to false positives) by 76\% and KLD (that penalizes false negatives) by 11\% overall, and achieves better results on some actions and context subsets (by up to 5--30\% KLD). This can be partially attributed to more meaningful spatio-temporal features extracted by the VST compared to other backbones.

We also investigate other techniques, such as weighted sampling (as in \cite{2018_ACCV_Xia}) and weighted loss. For both, we compute per-sample weight as a mean KLD score between sample ground truth and average saliency map for the video from which the sample is extracted. It has been hypothesized that samples that deviate from average contain important events \cite{2018_ACCV_Xia, 2018_PAMI_Palazzi, 2021_ICCV_Baee} and therefore choosing such samples more often during training or using the weight to penalize errors may be beneficial. Although both techniques lead to an additional overall improvement of 6\% KLD and 4\% NSS, high sample weights are not always correlated with drivers' actions and may instead amplify noise, leading to only minor improvements on some action and context subsets.

\begin{figure}[t!]
\centering
\includegraphics[width=\columnwidth]{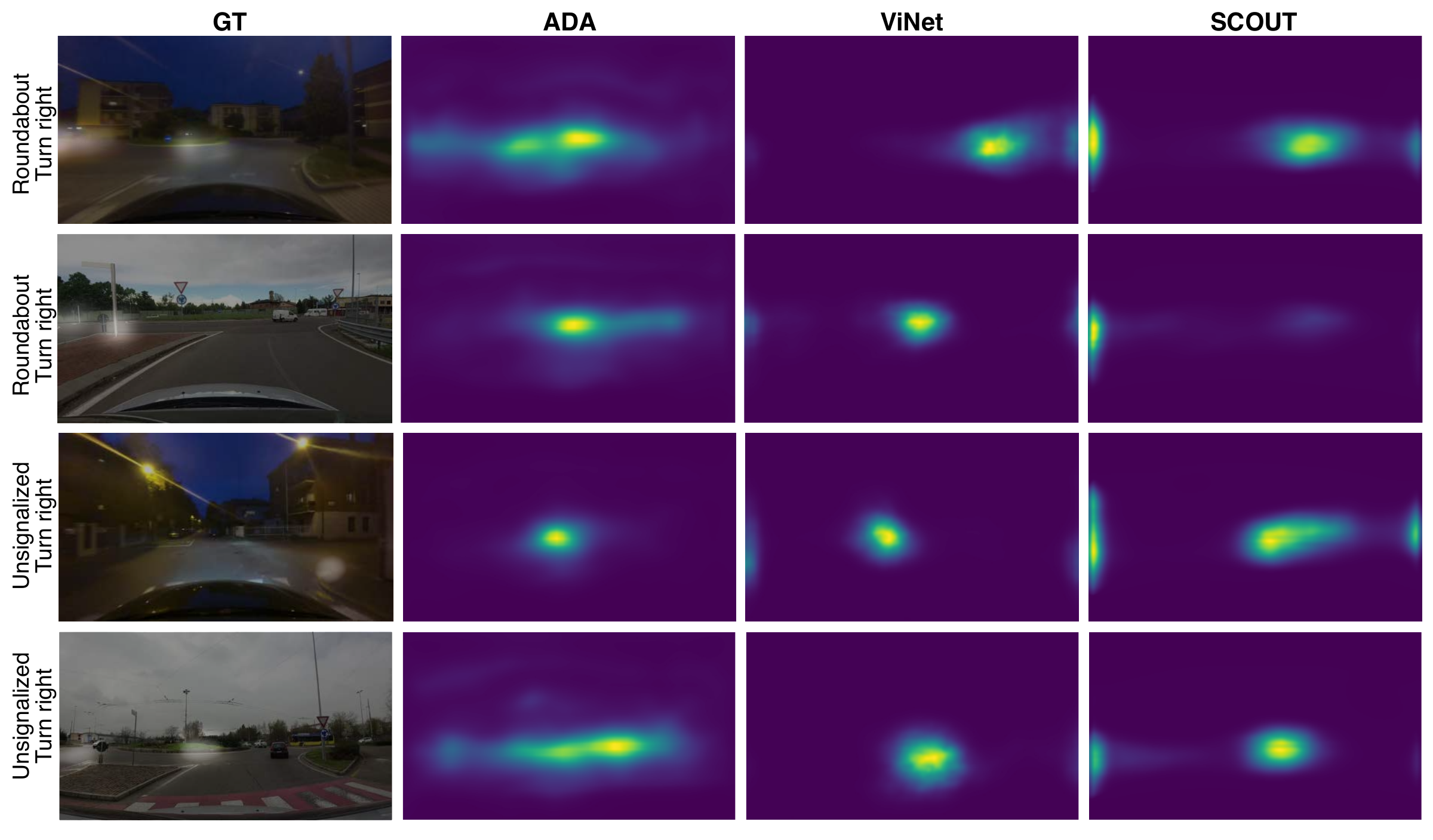}
\caption{Examples of yielding scenarios at different intersections types while performing a turn. The ground truth (in white) is overlaid on top of the images in the first column.}
\label{fig:qual_samples}
\vspace{-1.5em}
\end{figure}

\noindent
\textbf{Effects of task and context input.} Lastly, we experiment with task and context input to SCOUT: we vary the types of task/context features and whether they are incorporated into a single encoder block or several. The results in Table \ref{tab:SCOUT_results} indicate that any additional information is beneficial overall (89\% better on NSS and 24\% on KLD compared to SOTA, and 2--8\% across all metrics w.r.t. SCOUT w/o task) and especially for action and context subsets (10--30\% KLD over SOTA and SCOUT w/o task). The overall best performance is reached when task/context features are added to the second and third encoder blocks, presumably because they contain richer semantic information than earlier blocks and can take advantage of such representation. Significant performance gains (20--25\% KLD) are also achieved on the yielding scenarios at roundabouts and unsignalized intersections. This is demonstrated in Figure \ref{fig:qual_samples} rows 1, 2, and 4. Note that SCOUT accurately highlights the left side of the upcoming intersection where conflict vehicles may appear, whereas other models remain centered.  At the same time, some right-of-way subsets (roundabout and highway) see a slight decline in performance, caused by the model learning to always check the edges of the road near the intersections (see row 3 in Figure \ref{fig:qual_samples}) even if the ego-vehicle has the right-of-way.

%Model performance

%Results with weighted sampling - show that they are ineffective

%Results with different features 

%Results with addition of features at different places

%Results with different methods of mixing the task

%Qualitative results

\section{Conclusion}
In this paper, we discussed analyzing and modeling effects of task and context on drivers' gaze prediction. First, we extended the existing DR(eye)VE dataset with task- and context-related annotations and used them to demonstrate that most models do not capture gaze patterns typical for underrepresented and safety-critical scenarios, such as turning and yielding at intersections. We then showed via the proposed model that modulation with explicit task and context information is more effective than bottom-up learning or sampling strategies used in prior works. One limitation of the proposed model is reliance on manually annotated task labels due to noisy and incomplete vehicle information provided with the DR(eye)VE dataset (\eg inaccurate GPS and no yaw output useful for detecting lateral actions). In future work, we will explore methods for automatic extraction of these and other features representing context (\eg other road users) and other methods for modulating gaze prediction.

%%%%%%%%%%%%%%%%%%%%%%%%%%%%%%%%%%%%%%%%%%%%%%%%%%%%%%%%%%%%%%%%%%%%%%%%%%%%%%%%

%%%%%%%%%%%%%%%%%%%%%%%%%%%%%%%%%%%%%%%%%%%%%%%%%%%%%%%%%%%%%%%%%%%%%%%%%%%%%%%%

%%%%%%%%%%%%%%%%%%%%%%%%%%%%%%%%%%%%%%%%%%%%%%%%%%%%%%%%%%%%%%%%%%%%%%%%%%%%%%%%
%section*{Appendix 1}

\noindent
\textbf{Acknowledgement.} Work supported by the Air Force Office of Scientific Research under award number FA9550-22-1-0538 (Computational Cognition and Machine Intelligence, and Cognitive and Computational Neuroscience portfolios); the Canada Research Chairs Program (950-231659); Natural Sciences and Engineering Research Council of Canada (RGPIN-2022-04606).

%%%%%%%%%%%%%%%%%%%%%%%%%%%%%%%%%%%%%%%%%%%%%%%%%%%%%%%%%%%%%%%%%%%%%%%%%%%%%%%%

\bibliography{all}
\bibliographystyle{ieeetr} 

%\addtolength{\textheight}{-20cm}   % This command serves to balance the column lengths

%%%%%%%%%%%%%%%%%%%%%%%%%%%%%%%%%%%%%%%%%%%%%%%%%%%%%%%%%%%%%%%%%%%%%%%%%%%%%%%%
\appendix
\subsection{DR(eye)VE ground truth}
\label{suppl:gt}
\begin{figure*}[t!]
\centering
\includegraphics[width=\textwidth]{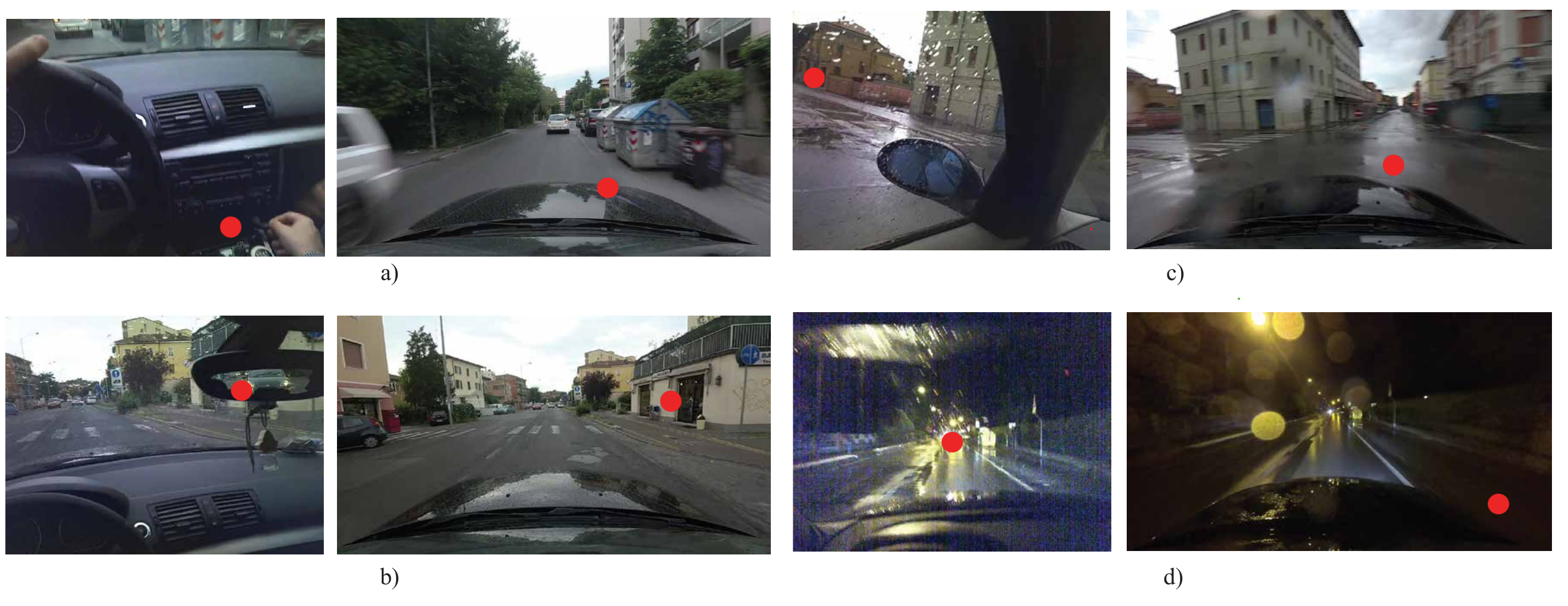}
\caption{Common scenarios where driver's gaze (shown as a red circle) is incorrectly projected onto the scene. Left column shows the view from the driver's eye-tracking glasses (ETG) and the right column shows the view of the rooftop Garmin camera (GAR). The following scenarios are demonstrated: a) looking down; b) checking rearview mirror; c) looking outside the GAR camera view during turns; d) raindrops on the windshield.}
\label{fig:projection}
\end{figure*}

Here we provide additional qualitative examples to illustrate some of the shortcomings of the original ground truth in DR(eye)VE dataset. Figure \ref{fig:projection} shows several common scenarios where fixations are incorrectly projected from the driver's eye-tracking glasses (ETG) view to the rooftop Garmin camera view (GAR): a) when the driver looks down, the scene is not visible in the ETG view and the homography cannot be established, thus driver's gaze is projected incorrectly onto the hood of the car; b) gaze towards the rearview mirror may be incorrectly projected onto elements of the scene behind the mirror; c) when making turns, the driver may look at the areas that are outside of the GAR camera view, however, these fixations are still mapped onto the scene in the original ground truth; d) since the GAR camera is mounted outside, the raindrops may cause issues with transformation even though the driver's gaze is within the GAR camera field of view.

\begin{figure*}[t!]
\centering
\includegraphics[width=\textwidth]{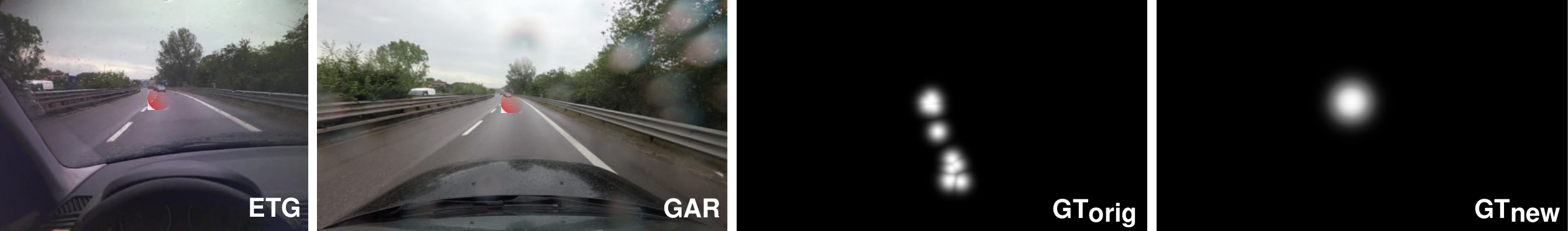}
\caption{Original (GT$_{orig}$) and new (GT$_{new}$) ground truth for the same frame (with fixations indicated with red circles in the drivers' (ETG) and rooftop camera (GAR) views). Note the ``trail'' of points in the original ground truth where the driver made a brief saccade towards the speedometer. The data points corresponding to saccaded are propagated across dozens of frames due to the aggregation procedure. These are absent in the new ground truth shown in the leftmost image.}
\label{fig:speedometer}
\end{figure*}

\begin{figure*}[t!]
\centering
\includegraphics[width=\textwidth]{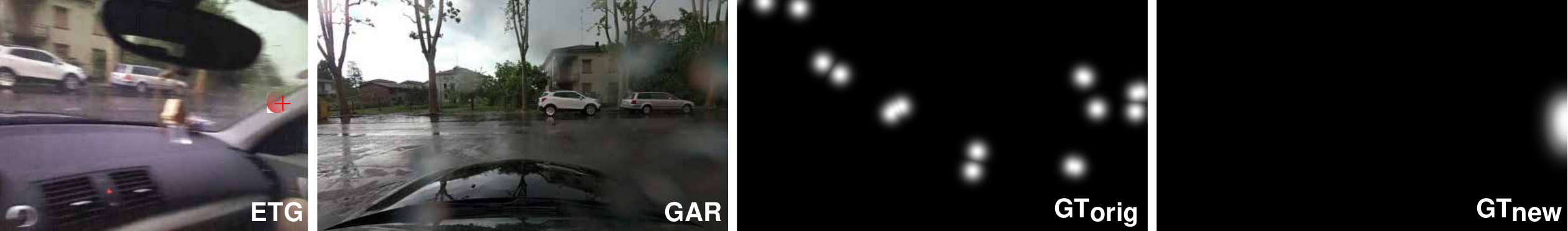}
\caption{Original (GT$_{orig}$) and new (GT$_{new}$) ground truth for the same frame (with fixations indicated with the red circle in the drivers' (ETG) view). The driver is looking through the side window, however, in the original ground truth, points representing saccades in the past frames and out-of-bounds fixations are still projected onto the scene in a random pattern. In the new ground truth, to preserve some gaze information, we push out-of-bounds fixations towards the edge of the frame to indicate the approximate direction of where the driver is attending.}
\label{fig:intersection}
\end{figure*}

Figure \ref{fig:speedometer} demonstrates the results of removing saccade to the speedometer from the ground truth. Figure \ref{fig:intersection} shows an example of ground truth containing driver's fixations that are outside of the GAR camera field of view.

\subsection{Task and context representation}
\label{suppl:task_context}
This section details how task and context labels were used for training the proposed SCOUT model.

Table \ref{tab:text_context_representation} lists all task and context information grouped into three sets: \textit{global context}---represents global weather, time of day, and location, \textit{local context}---information about the upcoming intersections and actions to be taken along the route, \textit{current action}---information about current longitudinal action (speed and acceleration) and lateral action (turns, lane changes, and driving straight).  

Global context labels were provided with the dataset and are unchanged. Local context information relies on the new manual annotations and GPS data provided with the dataset. The DR(eye)VE dataset description in \cite{2018_PAMI_Palazzi} does not specify whether the drivers knew the complete route in advance or were given step-by-step instructions. In the latter case, it is also unknown how well in advance such instructions were given and in what form. Therefore, we use guidelines designed for navigation systems that provide voice navigation commands to drivers \cite{campbell1998human}. We first compute the current distance in meters to the nearest intersection on the route using available GPS coordinates of the ego-vehicle. If that distance is larger than the max lead distance defined as $(\mathrm{speed} \mathrm{(km/h)}*2.22)+37.144$, we set it to \textit{inf} to indicate that the intersection is too far, otherwise, we use the distance directly.

To represent current longitudinal action, we use the sequence of speed and acceleration values. Raw speed values provided with the dataset are interpolated, passed through the median filter to remove outliers, and then used to compute acceleration. For lateral action, we use manually assigned text labels since heading angle is too coarse and inaccurate and yaw information is not available. If there is more than one label in the sample, the most frequently occurring label is chosen.

\begin{table}[t!]
\centering
\caption{Task and context representation}
\label{tab:text_context_representation}
\resizebox{\columnwidth}{!}{%
\begin{tabular}{@{}lll@{}}
\toprule
Category & Features & Values \\ \midrule
Global context & Weather & text label: sunny, cloudy, or rainy \\
 & Time of day & text label: morning, evening, night \\
 & Location & text label: highway, urban, suburban \\
Local context & Distance to intersection (m) & numeric value \\
 & Priority & right-of-way, yield \\
 & Next action & \begin{tabular}[c]{@{}l@{}}text label: turn right, turn left,\\ drive straight\end{tabular} \\
Current action & Speed (km/h) & numeric array \\
 & Acceleration (m/s2) & numeric array \\
 & Action & \begin{tabular}[c]{@{}l@{}}text label: turn right/left, lane change \\ right/left, drive straight\end{tabular} \\ \bottomrule
\end{tabular}%
}
\end{table}

%T. Sato M. Akamatsu, Analysis of drivers’ preparatory behaviour before turning at intersections

%Chuh-Fu Wu, A Study on the Design of Voice Navigation of Car Navigation System
%distance before even tdecision point should be speed * 2.5 seconds meters for the broadcasting to be finished.
%switching lanes is prohibited within 30 m from intersections in downtown areas, for major roads and expressways, the reaction distance should be speed * 2.5 s + 30 meters.

%Boll, Asif, Feel Your route: A tactile Display for Car navigation  https://ieeexplore.ieee.org/stamp/stamp.jsp?arnumber=5871576
%very far: 200-150m
%far: 100-150m
%near: 30-50m
%turn_now: 10m

%Alinaghi, Will You Take This Turn? Gaze-Based Turning Activity Recognition During Navigation https://drops.dagstuhl.de/opus/volltexte/2021/14764/pdf/LIPIcs-GIScience-2021-II-5.pdf
%Intersections of the world https://www.research-collection.ethz.ch/bitstream/handle/20.500.11850/283277/2/LIPIcs-GISCIENCE-2018-3.pdf

%https://dl.acm.org/doi/pdf/10.1145/2556288.2557404 

%HUMAN FACTORS DESIGN GUIDELINES FOR ADVANCED September 1998
%TRAVELER INFORMATION SYSTEMS (ATIS) AND COMMERCIAL
%VEHICLE OPERATIONS (CVO)
%US department of transportation
%minimum ideal distance = (speed (km/h) * 1.637) + 14.799
%ideal lead distance = (speed * 1.1973) + 21.307
%max lead distance = (speed*2.22) + 37.144

%Measuring Driver Perception: Combining Eye-Tracking and Automated Road Scene Perception
%https://journals.sagepub.com/doi/pdf/10.1177/0018720820959958

\subsection{Implementation of Kullback-Leibler divergence}
\label{suppl:KLD}
Kullback-Leibler divergence (KLD) is the key dissimilarity metric commonly reported when evaluating saliency algorithms \cite{2018_PAMI_Bylinskii}. 

%ADD CODE SAMPLES FROM THREE IMPLEMENTATIONS

%TODO: test UNISAL implementation

KLD is defined as $P\|Q = \sum p_i\mathrm{ln}\frac{p_i}{q_i}$, where $P$ and $Q$ are two distributions being compared. Since saliency maps are usually sparse, cases when non-zero value in one image ($p_i>0$) corresponds to zero value in another ($q_i=0$) occur frequently. To prevent division by zero, a small constant $\epsilon$ is added to the denominator in the equation above. Consequently, if the saliency values are small, the result can be dominated by this parameter. Depending on the implementation, $\epsilon$ may be set differently. Below are three typical values used: 

\begin{itemize}
\itemsep0em
\item MATLAB machine epsilon: $\epsilon=1.1920929\mathrm{e}-07$;
\item \texttt{numpy} machine epsilon: $\epsilon=2.2204\mathrm{e}-16$\footnote{\texttt{np.finfo(np.float32).eps}} (default in Python implementations);
\item A small constant: $\epsilon=0.0001$.
\end{itemize}

The Python implementation of KLD used in the DR(eye)VE evaluation code\footnote{\url{https://github.com/ndrplz/dreyeve/blob/master/experiments/metrics/metrics.py}} follows the widely used Matlab implementation \cite{bylinskii_metrics_code} for the MIT saliency benchmark\footnote{\url{saliency.mit.edu}}, but this Python version is numerically unstable for some inputs. We compared the DR(eye)VE version of KLD to two other Python implementations: Fahimi \& Bruce \cite{fahimi2021metrics, fahimi_metrics_code} and \texttt{pysaliency} \cite{mit-tuebingen-saliency-benchmark, kummererSaliencyBenchmarkingMade2018}, the official evaluation code for the MIT/Tuebingen saliency benchmark \footnote{\url{saliency.tuebingen.ai}}.

Table \ref{tab:KLD} shows results of testing different implementations with different values of $\epsilon$ on three test cases: 
\begin{enumerate}
\item \textit{Two identical randomly generated images}. KLD for two identical images is expected to be $0$, however, DR(eye)VE implementation fails to produce this value with MATLAB and small constant $\epsilon$ for two randomly generated identical images ($1000\times 1000$ px). 
\item \textit{Two different randomly generated images.} Similarly, DR(eye)VE implementation of KLD diverges significantly for MATLAB and small constant $\epsilon$ when tested on two \textit{different} randomly generated images. Two other implementations remain stable with small differences in the sixth decimal.
\item \textit{Randomly selected ground truth and predicted saliency maps from DR(eye)VE.} DR(eye)VE implementation diverges from others on a pair of ground truth and predicted saliency maps randomly selected from the dataset. Here, even using the default $\epsilon$ generates results different from other implementations.
\end{enumerate}

Table \ref{tab:dreyeve_results} shows the evaluation results of BDD-ANet \cite{2018_ACCV_Xia}, DReyeVENet \cite{2018_PAMI_Palazzi}, and ADA \cite{2022_T-ITS_Gan}, using DR(eye)VE implementation with MATLAB $\epsilon$ and Fahimi \& Bruce code with Numpy $\epsilon$. The results obtained with the DR(eye)VE implementation with MATLAB $\epsilon$ replicate the KLD values reported in the literature. Fahimi \& Bruce implementation produces different absolute values, but preserves the relative ranking of the models. Since finding the cause of the numerical issues is beyond the scope of these experiments, we report the results for all algorithms using the more accurate Fahimi \& Bruce implementation with Python $\epsilon$. 

\begin{table}[t!]
\centering
\caption{Results of testing different KLD implementations with several standard values of $\epsilon$ on three scenarios. Diverging results of DR(eye)VE implementation are highlighted in orange.}
\label{tab:KLD}
\resizebox{\columnwidth}{!}{%
\begin{tabular}{@{}lllll@{}}
 &  & \multicolumn{3}{c}{$\epsilon$} \\ \cmidrule(l){3-5} 
 &  & MATLAB & Numpy & Small constant \\
Data & Implementation & 1.19E-07 & 2.22E-16 & 0.0001 \\ \midrule
 & DR(eye)VE & \cellcolor[HTML]{FFCE93}-0.029013 & \cellcolor[HTML]{FFCE93}0 & \cellcolor[HTML]{FFCE93}-2.771208 \\
 & Fahimi \& Bruce & 0 & 0 & 0 \\
\multirow{-3}{*}{Identical images} & pysaliency & 0 & 0 & 0 \\ \midrule
 & DR(eye)VE & \cellcolor[HTML]{FFCE93}0.087539 & \cellcolor[HTML]{FFCE93}0.499305 & \cellcolor[HTML]{FFCE93}-5.205599 \\
 & Fahimi \& Bruce & 0.499304 & 0.499305 & 0.498398 \\
\multirow{-3}{*}{Random images} & pysaliency & 0.499304 & 0.499306 & 0.498398 \\ \midrule
 & DR(eye)VE & \cellcolor[HTML]{FFCE93}2.322985 & \cellcolor[HTML]{FFCE93}7.277252 & \cellcolor[HTML]{FFCE93}-2.70738 \\
 & Fahimi \& Bruce & 5.135229 & 9.888739 & 3.525612 \\
\multirow{-3}{*}{Salmaps from DR(eye)VE} & pysaliency & 5.135229 & 9.888739 & 3.525612 \\ \bottomrule
\end{tabular}%
}
\end{table}

\begin{table}[t!]
\centering
\caption{Evaluation results on the original DR(eye)VE dataset ground truth using DR(eye)VE implementation with MATLAB $\epsilon$ and Fahimi \& Bruce code with Numpy $\epsilon$.}
\label{tab:dreyeve_results}
\resizebox{\columnwidth}{!}{%
\begin{tabular}{@{}ccc@{}}
\toprule
Model & KLD (DR(eye)VE implementation) & KLD (Fahimi \& Bruce) \\ \midrule
BDD-ANet & 1.92 & 3.15 \\
DReyeVENet & 1.59 & 2.63 \\
ADA & 1.56 & 2.33 \\ \bottomrule
\end{tabular}%
}
\end{table}

\subsection{Additional benchmark results}
\label{suppl:benchmark_results}

In the paper we reported model results on action and context sequences using only the KLD metric. The reason for that decision was that majority of models (including our own) were trained with KLD loss since empirically it has been shown to result in the best performance across all metrics. Here, for completeness we provide additional evaluation results on CC, SIM, and NSS metrics. 

Additional benchmark results on DR(eye)VE in Table \ref{tab:benchmark_extra} demonstrate consistent effects of action and context on the performance of the bottom-up models of drivers gaze. Specifically, actions involving lateral maneuvers as well as yielding at roundabouts and unsignalized intersections are the most challenging for all models.  

Table \ref{tab:SCOUT_extra} shows that the proposed SCOUT model achieves competitive results on the most challenging scenarios (in DReyeVE and BDD-A) with respect to other metrics as well.

\begin{table*}[t!]
\centering
\caption[Additional benchmark results on DR(eye)VE with respect to CC, NSS, and SIM metrics]{Additional benchmark results on the \textbf{new} DR(eye)VE ground truth with respect to CC, NSS, and SIM metrics.$\uparrow$ indicates that larger values are better. Green and red colors indicate better and worse performance, respectively. Best and second-best values are shown as \textbf{bold} and \underline{underlined}. Abbreviations: None---maintain speed/lane, Dec---longitudinal deceleration, Acc---longitudinal acceleration, Lat---lateral actions only, Lat/Lon--simultaneous lateral and longitudinal actions, Stop---ego-vehicle is stopped, RoW---ego-vehicle has right-of-way.}
\includegraphics[width=\textwidth]{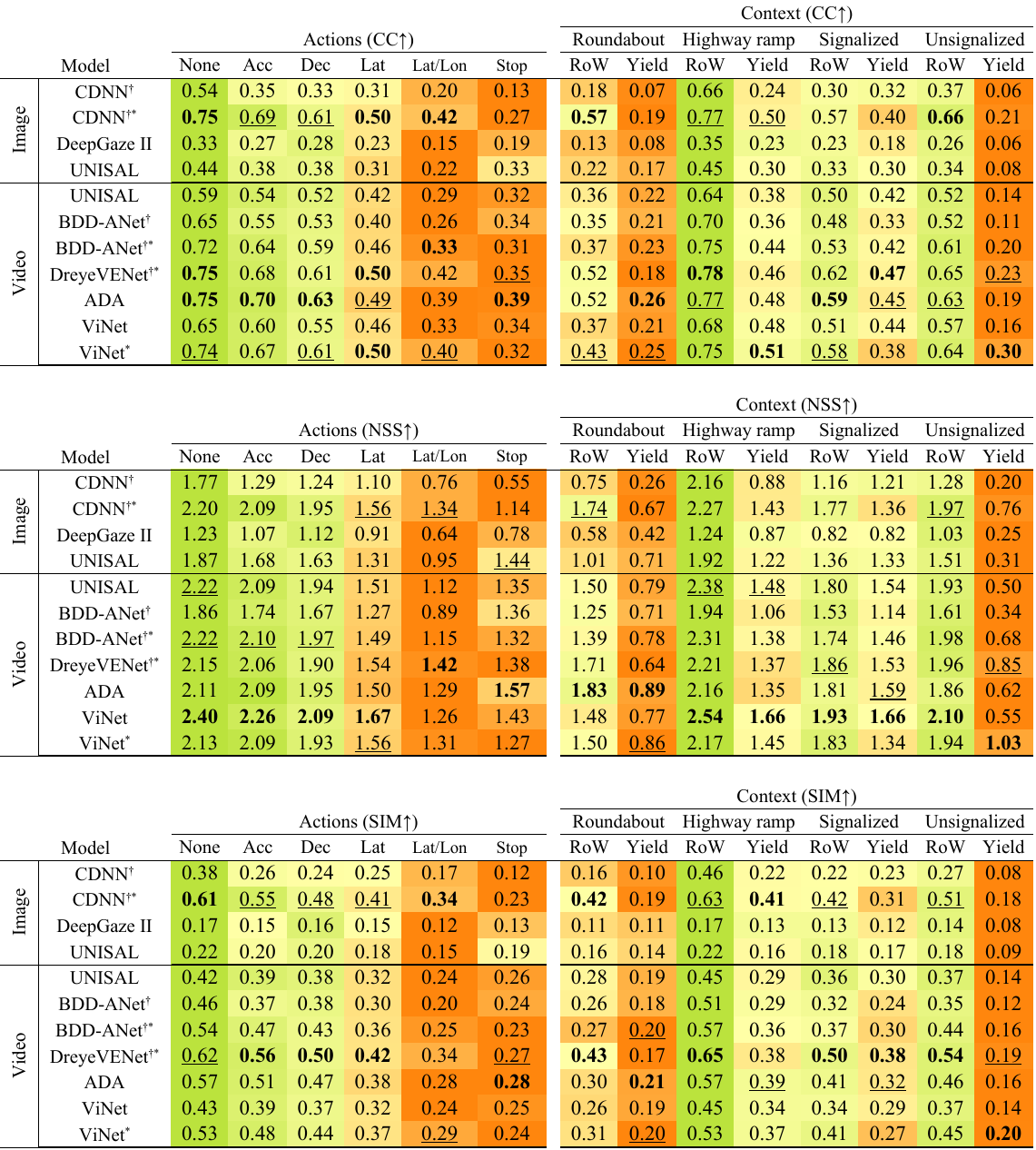}
\begin{flushleft}
\begin{footnotesize} *---model is  trained on the data; \textdagger ---model for drivers' gaze prediction.
\end{footnotesize}
\end{flushleft}
\label{tab:benchmark_extra}
\end{table*}

%\chapter{Additional SCOUT results}
%\label{suppl:SCOUT_results}
%
%Table \ref{tab:SCOUT_extra} reports additional evaluation results on action and context sequences and CC, SIM, and NSS metrics for variants of the proposed SCOUT model.

\begin{table*}[t!]
\centering
\caption[Additional SCOUT results on DR(eye)VE with respect to CC, NSS, and SIM metrics]{Additional results for variants of the SCOUT model on the \textbf{new} DR(eye)VE ground truth with respect to CC, NSS, and SIM metrics. $\uparrow$ indicates that larger values are better. Best and second-best values are shown as \textbf{bold} and \underline{underlined}. Abbreviations: None---maintain speed/lane, Dec---longitudinal deceleration, Acc---longitudinal acceleration, Lat---lateral actions only, Lat/Lon--simultaneous lateral and longitudinal actions, Stop---ego-vehicle is stopped, RoW---ego-vehicle has right-of-way.}
\includegraphics[width=\textwidth]{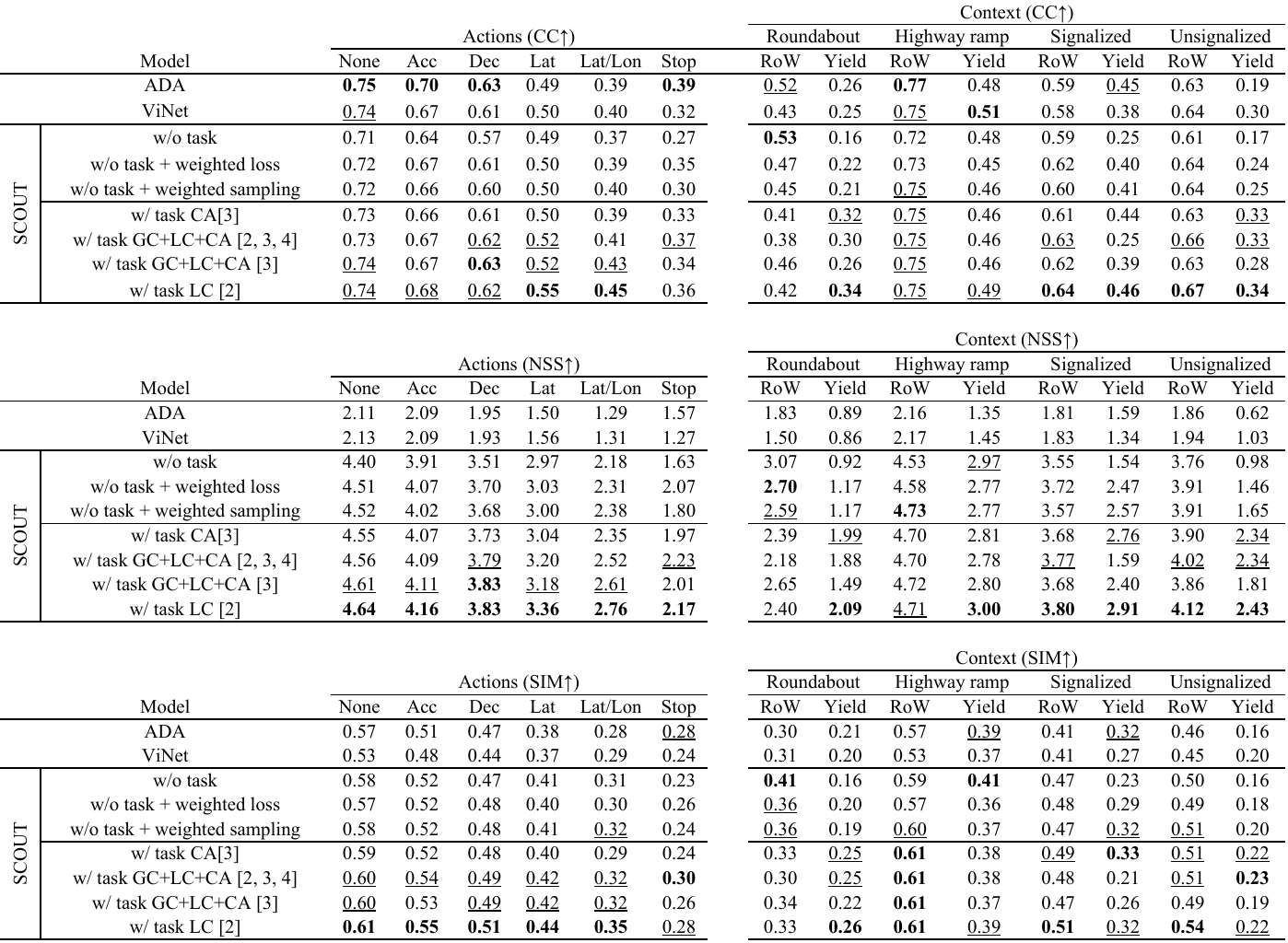}
\begin{flushleft}
\begin{footnotesize} *---model is  trained on the data; \textdagger ---model for drivers' gaze prediction.
\end{footnotesize}
\end{flushleft}
\label{tab:SCOUT_extra}
\end{table*}

\end{document}